\documentclass[letterpaper]{article}
\usepackage{aaai17}
\usepackage{graphicx}

\title{Practical Challenges in Explicit Ethical Machine
  Reasoning\thanks{The work in this paper was supported by the EPSRC
    ``Verifiable Autonomy'' project (EP/L024845)}} 
\author{Louise A. Dennis \and Michael Fisher \\ 
        Department of Computer Science \\
        University of Liverpool, UK \\ 
        \texttt{L.A.Dennis@liverpool.ac.uk}
        \and \texttt{MFisher@liverpool.ac.uk}} \nocopyright

\newcommand{\geneth}{{\sc GenEth}}
\newcommand{\ethan}{{\sc Ethan}}

\begin{document}
\maketitle

\begin{abstract}
We examine implemented systems for ethical machine reasoning with a
view to identifying the practical challenges (as opposed to
philosophical challenges) posed by the area.  We identify a need for
complex ethical machine reasoning not only to be multi-objective,
proactive, and scrutable but that it must draw on heterogeneous
evidential reasoning.  We also argue that, in many cases, it needs to
operate in real time and be verifiable.  We propose a general
architecture involving a declarative ethical arbiter which draws upon
multiple evidential reasoners each responsible for a particular
ethical feature of the system's environment.  We claim that this
architecture enables some separation of concerns among the practical
challenges that ethical machine reasoning poses.
\end{abstract}

\section{Introduction}

There has been an explosion of interest in Ethics and Artificial
Intelligence as evidenced by several high profile initiatives
considering the issue such as the \emph{IEEE Global Initiative on
  Ethics in Artificial Intelligence and Autonomous Systems} and the
\emph{BSI Standard 8611: Guide to the Ethical Design and Application
  of Robots and Robotic Systems}.  While these initiatives generally
take a wide-ranging view of the subject considering everything from
the deployment of autonomous weapons, the societal impact from the
potential loss of jobs, to the privacy issues that result from big
data and social media they also consider, as a topic, the
implementation of ethical reasoning in machines, often referred to as
machine ethics but which we will here refer to as ethical machine
reasoning in order to highlight our consideration of computational
reasoning about ethical issues.

One of the key challenges facing the implementation of ethical machine reasoning is that no consensus exists on the nature of morality, the key moral values, how morals relate to ethical rules and how competing ethical rules can be decided between in specific contexts.   We will refer to these issues as \emph{philosophical challenges} facing the implementation of ethical machine reasoning.

In this paper we contend that there are a range of other challenges
faced by ethical machine reasoning which would make it a challenging
area of artificial intelligence \emph{even if} the philosophical
challenges were resolved.  These \emph{practical challenges} relate to
questions of how ethical reasoning is to be implemented.  They let us
identify the implementation of ethical machine reasoning as a distinct
sub-field of automated reasoning in general and demonstrate that it is
not possible to satisfactorily implement ethical machine reasoning
simply by taking pre-existing automated reasoning techniques and
applying them to the ethical theory of your choice.

In this paper we seek to understand the practical challenges that
characterise machine ethics.  We frame this understanding around the
discussion of existing systems that claim to implement ethical
reasoning and propose a general software architecture for ethical
machine reasoning which would support a variety of solutions to these
challenges.

\section{Survey of Ethical Machine Reasoning Implementations}

All machine reasoning systems can be viewed a ethical reasoning
systems from some level of abstraction, so we here restrict ourselves
to systems that are explicitly ethical in the sense of
Moor~\cite{Moor06} (i.e., they reason explicitly about ethical
concepts).  There are few examples of such systems and we view the
main ones here -- our purpose in doing so is to highlight the
practical issues faced by the implementation of ethical machine
reasoning.

\subsection{Ethical Governors}

The first implementation of ethical machine reasoning is generally
credited to Arkin et. al, of the Georgia Tech Mobile Robot
Lab~\cite{Arkin09,Arkin12} who outline the architecture for an
\emph{ethical governor} for automated targeting systems for autonomous
weapons.  This governor was charged with ensuring that any use of
lethal force was governed by the ``Law of War'', the ``Rules of
Engagement'' and was \emph{proportional}.

The governor was implemented as a separate module that intercepted
signals from the underlying deliberative system and, where these
signals involved lethality, would go through a process of
\emph{evidential reasoning} which amassed information about the
situation in a logical form and would then reason about the evidence
using constraints represented as prohibitions and obligations.  If any
prohibitions were violated or obligations unfulfilled then the
proposed action would be vetoed.  If no prohibition were violated then
the governor would proceed to a ``collateral damage'' estimation phase
and attempt to find a combination of weapon system, targeting pattern
and release position that would maximise the likelihood of
neutralising the target while minimising collateral damage.

The authors note that ``it is a major assumption of this research that
accurate target discrimination with associated uncertainty measures
can be achieved despite the fog of war" and the case studies they
perform using their implementation are based on this assumption and
provide appropriate information up front as part of the scenario.

This initial work on ethical governors was then re-implemented in a
new setting of healthcare~\cite{Shim17}.  In this setting the ethical
governor monitors not an underlying autonomous system but the
interactions between a patient with Parkinson's Disease and a
caregiver.  The reasoning behind such a monitoring system is that
sufferers from Parkinson's Disease frequently lose control of their
facial musculature, this means that many non-verbal cues and
interactions between patient and care-giver are lost which can result
in stigmatisation between a caregiver and patient and a decrease in
the quality of patient care.  This ethical governor combines an
evidential reasoner with a rule-based system (as opposed to a
constraint reasoning system) but the basic architecture is the same.
The evidential reasoner produces an assessment of the environment
based on cues such as raised voices which are represented in a logical
form.  The rule-based system then reasons about this logical
information in order to select appropriate intervening actions such as
verbal interventions or indicative gestures.

\subsection{{\sc GenEth}}
The \geneth\ system~\cite{Anderson14} is designed as an ethical
dilemma analyzer.  Its purpose is twofold. Firstly, it demonstrates
how input from professional ethicists can be used via a process of
machine learning to create a \emph{principle of ethical action
  preference}\footnote{We note that the terminology of ethical
  principles is used widely but inconsistently throughout the
  literature.} which can be applied to situations in order to
determine appropriate action.  \geneth\ analyses a given situation in
order to determine its ethical features (e.g., that harm may befall a
patient in some healthcare scenario), these features then give rise to
duties (to minimize or maximize that feature).  The principle of
ethical action preference is used to compare two options: each option
is assigned a score for each relevant ethical feature, the difference
between these two scores is then used by the principle which
partitions the $n$-dimensional space defined by the feature
comparisons (one dimension for each feature) into regions using
inequalities.  Each partition of the space specifies which of the
compared actions is to be preferred in that region so, for instance if
the first action is significantly worse in terms of privacy than the
second (e.g., the difference in their score on the privacy feature is
greater than 2) but the second action is a little worse in terms of
patient safety (e.g., the difference in their score is less than -1)
then the partition might specify that the first action is to be
preferred.

Initially \geneth\ was implemented as a standalone system which was used to capture information from medical ethicists on decisions that should be made in particular, manually generated, scenarios.  It has subsequently been connected as a decision-making component on top of a simulator for Nao robots~\cite{Anderson16} and evaluated in scenarios where the robot must choose between six possible actions (such as charging itself, reminding a patient to take medication, and notifying an overseer of problems).  These actions are evaluated on an ongoing basis using a principle which considers eight ethical features (honour commitments, maintain readiness, minimise harm, maximise good, minimise non-interaction, respect autonomy and maximise the prevention of immobility).

\geneth\ is able to give explanations for its decisions in terms of its partition of the space -- so it can state how two options compared on the various ethical features in its judgement and refer to the statement of the principle to then justify the subsequent choice.

\subsection{Ethical Consequence Engines}
Winfield et. al, of the Bristol Robotics
Lab~\cite{Winfield13,Vanderelst16} have investigated systems based on
the concept of an \emph{Ethical Consequence Engine}.  This consequence
engine uses simulation to evaluate the impact of actions on the
environment.  In particular it simulates \emph{not just} the actions
of the robot itself but also simulates the activity of other agents in
the environment, based on some simplifying assumptions about movement
and intended destinations of humans and other robots.  This allows the
robot to determine not only if its actions have directly negative
consequences (e.g., colliding with a person) but if they have
indirectly negative consequences (e.g., failing to intercept a person
who might otherwise come into danger).

There are two versions of the ethical consequence engine system the first of which~\cite{Winfield13} was implemented on e-Puck robots and evaluated all possible actions in the environment based on a discritization of the space of operation, while the second~\cite{Vanderelst16} was implemented on Nao robots and used a sampling procedure to evaluate a specific sub-set of options.  Each option is scored using various metrics such as the closeness of any humans to ``danger", the closeness of the robot to ``danger" and the closeness of the robot to its goal.  These metrics are then combined in a weighted sum and the highest scoring option chosen.  The first of these systems was also verified~\cite{Dennis15} in a process that involved converting the metric based evaluation of the robot actions into logic based reasoning over outcomes that compared the severity of the outcome and who suffered the concequences (human or robot).

\subsection{{\sc Ethan}}
The \ethan\ system~\cite{Dennis16} was developed to investigate ethical decision making in exceptional circumstances with a particular emphasis on verifiability.  In the \ethan\ system a \emph{rational agent}, based on the \emph{Beliefs-Desires-Intentions} model of agency~\cite{Rao95} was used to reason about the ethical risks of plans proposed by an underlying planning system.  In this system the operation of reasoning in normal circumstances was assumed to be ethical by default (i.e., there was an assumption that appropriate ethical properties were guaranteed by a process of testing or verification of the decision-making process) but that in exceptional circumstances the system might need to make use of Artificial Intelligence techniques such as planning or learning which are inherently challenging to verification.  \cite{Dennis16} considers the case of a planning system that returns candidate plans to the agent which are annotated with any ethical principles\footnote{These can be considered broadly equivalent to \geneth's ethical features.} impacted by the plan.  The case study looked at scenarios involving unmanned aircraft and plans were annotated with the nature of any collisions that might take place or violations of the Rules of the Air.  \ethan\ then reasoned using a context specific \emph{ethical policy} which imposed an ordering on plans based upon the ethical principles they violated.  

Model Checking~\cite{Clarke99} was then applied to the \ethan\ agent in order to verify a number of properties, including that the agent was programmed to correctly obey the specified ethical policy -- i.e., that if it selected a plan that violated some ethical principle then this was only because all the other available options were worse.

\section{Observed Features}

We can observe a number of features both individually and jointly
across these systems that begin to define the space of practical
challenges that face systems seeking to implement explicit ethical
reasoning.



\subsection{Multi-Objective}

While terminology across these systems is inconsistent we note that all of them operate on the assumption that the situation in which the system finds itself may have a number of potential ethical impacts -- whether these are referred to as ethical features, ethical principles, ethical constraints or by some other language.  We will refer to these as ethical features for convenience.  

While we anticipate that in most everyday reasoning situations at most
one ethical feature is at stake -- i.e., in many cases none of the
available options have particular ethical features (\emph{answering
  the front door}, for instance) -- in nearly all cases the point of
the ethical reasoning is to limit goal-directed behaviour according to
ethical considerations (though some of the systems treat goal-directed behaviour as an ethical feature expressed, for instance, as obedience to the human).  However, consideration of ethical features
rapidly leads to the conclusion that there will be situations where
the system must somehow choose between them, as well as limiting
goal-directed behaviour.  So \geneth\ has its partitioning of the
space that compares individual options according to their ethical
features, \ethan\ has its context-dependent ethical policy while the
Ethical Consequence Engine prioritises humans over robots, and within
that the extent of the harm that may befall the agent.

We note that this makes ethical reasoning {\bf inherently
  multi-objective} which is a practical challenge to many techniques
for controlling decision-making in machines.  In particular this kind
of reasoning is challenging for techniques that seek to maximize or
minimize some value for while at a very abstract level we can say the
point of ethical reasoning is to maximise human well-being (as
suggested by~\cite{Dignum18}) this is not a concept easily captured in
a function.  Instead abstract concepts such as well-being are
concretized as ethical features -- ethical machine reasoning then
becomes about deciding what balance among these features is most
likely to improve or preserve human well-being (or some equivalent
abstract general value)\footnote{We probably need to accept that in
  the absence of an agreed philosophical framework for morality, the
  best any feature-based reasoning can hope to achieve is choosing the
  right outcome most of the time.}.

It is tempting to drop down to a lower level and utilise
multi-objective optimisation~\cite{Miettinen98}, but this works
against many other features we require such as scrutability and
verifiability.

\subsection{Heterogenous Evidential Reasoning}

The Ethical Governor systems are the only approaches that explicitly
describe their architecture as consisting of first an evidential
reasoner which translates sensory information into a logical form and
then a second reasoner that makes a decision based upon the logical
translation.  However both \geneth\ and \ethan\ also take something
approaching this form, assuming that information has been expressed in
some logical or equational form for use by the system.  Interestingly
the case studies presented for all these systems either adopt very
simple evidential reasoning mechanisms, or use some oracle to provide
information in an appropriate form and focus on the subsequent
reasoning.

While the ethical consequence engines do not use a logical expression of data, they too crucially involve an evidential reasoning phase by using a simulator to make predictions about the outcomes of actions.  It is also clear, particularly in~\cite{Winfield13} that although the simulation results are converted to metrics and then employed in a utility function it is envisaged that this captures logical-style reasoning about the severity of outcomes and the relative importance of humans and robots.

The ethical consequence engines use simulation to make predications about safety outcomes, but it is easy to see that simulation is not effective in, for instance, establishing risks to human free will and autonomy and while simulations of information flows might be sufficient to determine privacy risks in social media settings, it is unlikely to be sufficient when considering information flow around smaller groups of people such as families and health workers.  An ethical machine reasoning system operating on a complex set of ethical features will need to use a {\bf variety of heterogeneous mechanisms to perform evidential reasoning} about the situation it finds itself in.  

The nature of this heterogeneous evidential reasoning appears to be a particularly under-explored aspect of ethical machine reasoning, even given the relative youth of the field and the small number of implemented systems.  It is also of note that the ethical features considered by the ethical systems we survey vary wildly, sometimes within a given system -- for instance the \geneth\ case study~\cite{Anderson16} considers both ``readiness" (which relates primarily to how much charge the robot has) and the far more abstract concept of ``good'' as ethical features.  \ethan\ treats ``Do not collide with people'' and ``Do not collide with aircraft'' as distinct ethical features despite the fact that both are clearly related to safety.  Understanding of what makes a suitable ethical feature, as an atomic concept for ethical machine reasoning and whether there is some heirarchy among these (e.g., safety is associated with features specifying \emph{whose safety}), and the extent to which they need to be annotated with, for instance the degree of severity of the impact and the uncertainty  about the outcome, is lacking and is a challenge that clearly has both practical and philosophical aspects.

\subsection{Real Time}
Most of these systems have had to make some compromise with the real time aspects of ethical reasoning in machines.  We note that this is not always the case.  The ethical governor system that mediates between patient and care-giver has more time available for reasoning than does the ethical consequence engine must react quickly to prevent an accident and in general we can envisage advisory systems for committees of people that would have minutes rather than fractions of a second for deliberation.

However, as a general observation, ethical machine reasoning must often perform {\bf complex reasoning in real time}.

\subsection{Proactivity}

Most of the implementations we have surveyed are \emph{reactive} -- i.e., their purpose is to veto or order plans/actions suggested by the underlying system.  However several of them acknowledge a need for {\bf proactivity} -- the ability not only to veto plans but to suggest separate courses of action.  This is most obvious in the Ethical Consequence Engine in which the whole point of the experiment is to divert the robot away from its goal-directed task for ethical reasons, but also in the intervening ethical governor which does nothing unless ethical considerations prompt an intervention.  

In the initial implementation of the ethical consequence engine~\cite{Winfield13} the underlying engine suggested all possible alternatives to the governor via a discretization process.  However the later version~\cite{Vanderelst16} presented only a  limited number of options for practical reasons -- these options were generated by the underlying control system which was therefore clearly implicitly using ethical considerations in order to generate options for the ethical layer.  Researchers at Bristol Robotics Lab are now seeking to have the ethical layer request options itself if those generated by the underlying controller are deemed insufficiently ethical\footnote{Paul Bremner, personal communication.} and this would be the first concrete implementation of ethical proactivity in such systems.

More generally, particularly in cases where the ethical principle of human free will is concerned, it may be necessary for an ethical reasoner to go through an information discovery process in order to determine the wishes of the persons it is interacting with and the strength of those wishes before deciding whether or not it is ethical to intervene.  For instance, it is a well established principle that people should be allowed to smoke in their own homes but considerable societal resource is put into making sure they are aware of the risks associated with smoking.  We might want a home-support robot to confirm that a home-owner was aware of the dangers of smoking but, if they were, to thereafter allow them to smoke without intervention.

\subsection{Scrutability}

In general it is desirable for a number of reasons that we should be able to understand how a machine reasons: for instance in order to predict its behaviour, and diagnose errors.  However this is particular important where ethical reasoning is concerned.   Indeed Moor's classification of ethical machine agents specifies that explicit agents, such as we consider here, should be able to justify their choices~\cite{Moor06}.

While moral philosophy has reached no consensus about whether morality is absolute or relative and societally determined, we view ethical machine reasoning as reasoning about circumstances where a systems actions may impact on the values of a community or individual.  Moor refers to these as ethical impact agents.  If some concept has attained the status of a moral value, then it has assumed a place of critical importance.  Therefore, while people may accept (even if they are irritated) that their SatNav sometimes chooses strange routes without explanation, they are less likely to accept that some system has chosen to violate their privacy and can offer no explanation for why it did so.  The ability to understand how a system has reached some decision is variously referred to as {\bf explainability} or {\bf scrutability}.

There are a number of forms scrutability~\cite{CaminadaKOV14} can
take, from being able to inspect design/requirements documents that
set out the rules of ethical behaviour the robot is following to the
ability to extract an explanation from the robot after a decision that
has been made to justify that decision or reconstruct the decision-making process after some problem arises (e.g., an \emph{ethical black box}~\cite{Winfield17}).

We see this need for scrutability in a number of the systems we survey.  Both the ethical governors and \geneth\ can explicitly justify their reasoning and the use of BDI style agents in \ethan\ also points to this concern, since their reasoning is based upon logic programming, which in turn is based upon logical deduction using explicit rules.  While such derivations can be complex to follow they are designed to mimic an account of human reasoning.

\subsection{Verifiability}

A key value in (arguably) all human societies is human well-being, which manifests ethically as considerations around human safety (among other things).  Where a system is deemed to be \emph{safety critical} we are used to requiring high standards of verification.   We would argue that these considerations extend to \emph{ethically critical} systems. If a value is of sufficient importance to a community to be considered a moral value then we should expect {\bf high standards of verification} for any system with the potential to have a serious negative impact on that value.

Since, as we have noted, it is beyond our capability to simply insist autonomous systems maximise human well-being we are instead forced to encode ethical reasoning as a set of rules and these sets of rules are likely to be complex -- for instance even the fairly simple case study examined in~\cite{Anderson16} partitions the space of options into 13 regions, each representing a different combination of the ethical features of a situation.  There is therefore scope for error both in implementing a stated set of rules into a machine and expressing the rules so that they do indeed reflect our values.  Both~\cite{Dennis15} and~\cite{Dennis16} consider formal properties for machine ethics systems that can be checked by model-checking so long as the system itself has been implemented with such verifiability in mind.

\section{Architecture:  Multiple Ethical Governors}

\begin{figure}[htpb]
\centering
\includegraphics[width=0.4\textwidth]{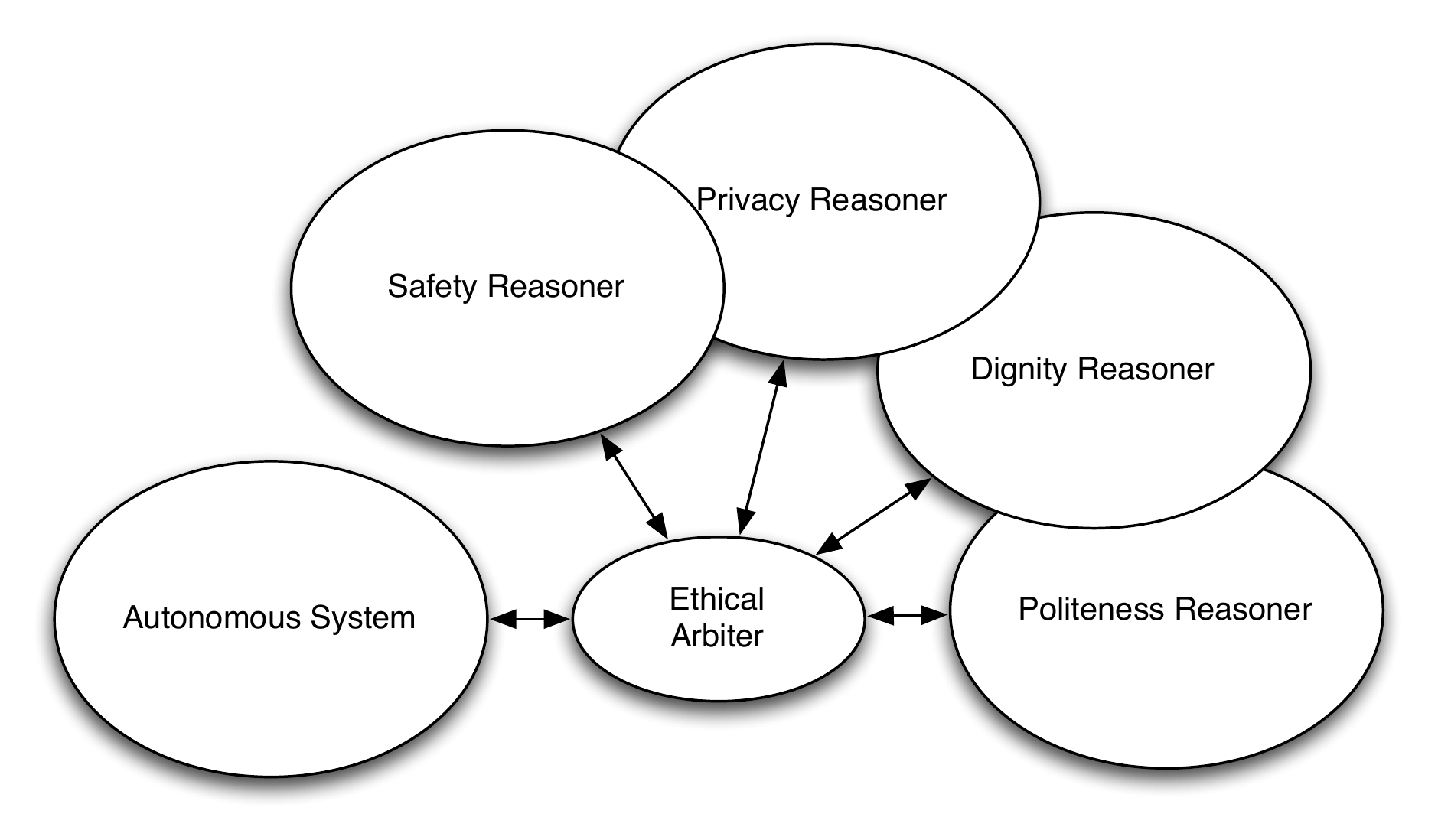}
\caption{An Architecture for Ethical Machine Reasoning}
\label{fig:architecture}
\end{figure}

We propose a generic architecture for ethical reasoning shown in Figure~\ref{fig:architecture}.  In this architecture an \emph{ethical arbiter} reasons using evidence provided by a number of \emph{evidential reasoners} each of which is customised to reason appropriately about some particular ethical feature of the domain.  The underlying autonomous system communicates its options to the ethical arbiter, these options are assessed by the evidential reasoners which convert information about the options into a logical or equational form which the ethical arbiter then reasons over.  The ethical arbiter then communicates the result of this reasoning back to the autonomous system.
The arbiter itself should be declarative in nature (i.e. the programming should focus on expressing the logic of the computation as opposed to its control flow).   Declarative programming supports scrutability at design time (since the program itself should focus on the outcomes of execution as opposed to how those outcomes are generated) and declarative programming paradigms in general also have better support for verification.  Logic programs, BDI agents and constraint reasoners are all examples of declarative programs.

One advantage we claim for this architecture is it allows us to
allocate some of the challenges faced by ethical machine reasoning to
different parts of the system.  Scrutability and multi-objective
reasoning are the preserve of the ethical arbiter, while real time
concerns can be partitioned into those requiring real time evaluation of the situation (which is the concern of the evidential reasoners) and efficient declarative reasoning (the concern of the arbiter).  The modularity
also gives us the potential to verify the ethical reasoning itself
separately (e.g., following the methodology in~\cite{Dennis14}) from
any verification of the accuracy of the evidential reasoners.

In Figure~\ref{fig:architecture}, we have included a ``politeness
reasoner''.  This is because many of the considerations that apply to
ethical machine reasoning we believe also apply to machine reasoning
about social norms -- in particular that such reasoning is
multi-objective and needs to be proactive.  It is therefore possible
to imagine such an architecture being extended to cover more general
normative reasoning as well as specifically ethical reasoning.

Many design choices exist within this architecture such as whether the evidential reasoners and/or the arbiter can suggest or request new actions/plans; the nature of the evidence produced (which could potentially contain information pertaining to certainty, severity, who is impacted, how many people are impacted and so on); what constitutes an atomic ethical feature that grounds out reasoning; and so on.   The architecture also allows rich or sparse logics to be used by the arbiter.

\section{Conclusion}

We here address the challenges posed by explicit ethical machine
reasoning that are not related specifically to the philosophical
uncertainty surrounding the subject matter.

We have argued that explicit ethical machine reasoning faces challenges relating to its multi-objective nature, its frequent requirement for real time process, the heterogenous nature of the evidence it needs to reason about and challenges relating to its scrutability and verifiability.  Taken together we believe these challenges make ethical machine reasoning a sub-field of interest not just to philosophers and those interested in formal reasoning about ethics but also to those interested in the implementation of machine reasoning in general.

We have proposed a generic architecture (see
also~\cite{DennisFisher18:AIES}) which we consider suitable for
explicit ethical machine reasoning which, among other things,
modularises the ethical reasoning component and so allows some
separation of the various challenges into distinct sub-systems,
providing routes for tackling these problems.  None of the challenges
highlighted in this paper are solved by the architecture but it is our
intention in future work to implement the architecture and use it as a
vehicle for tackling the various practical challenges posed by
explicit ethical machine reasoning.


\begin{thebibliography}{}

\bibitem[\protect\citeauthoryear{Anderson and Anderson}{2014}]{Anderson14}
Anderson, M., and Anderson, S.~L.
\newblock 2014.
\newblock Geneth: A general ethical dilemma analyzer.
\newblock In {\em Proceedings of the Twenty-Eighth AAAI Conference on
  Artificial Intelligence}, AAAI'14,  253--261.
\newblock AAAI Press.

\bibitem[\protect\citeauthoryear{Anderson, Anderson, and
  Berenz}{2016}]{Anderson16}
Anderson, M.; Anderson, S.~L.; and Berenz, V.
\newblock 2016.
\newblock {A Value Driven Agent : Instantiation of a Case- Supported
  Principle-Based Behavior Paradigm A Value Driven Agent : Instantiation of a
  Case-Supported Principle-Based Behavior Paradigm}.
\newblock In {\em AAAI 2016 Workshop on AI, Ethics {\&} Society}.
\newblock AAAI Press.

\bibitem[\protect\citeauthoryear{Arkin, Ulam, and Duncan}{2009}]{Arkin09}
Arkin, R.; Ulam, P.; and Duncan, B.
\newblock 2009.
\newblock {An Ethical Governor for Constraining Lethal Action in an Autonomous
  System}.
\newblock Technical Report GIT-GVU-09-02, Mobile Robot Laboratory, College of
  Computing, Georgia Institute of Technology.

\bibitem[\protect\citeauthoryear{Arkin, Ulam, and Wagner}{2012}]{Arkin12}
Arkin, R.; Ulam, P.; and Wagner, A.
\newblock 2012.
\newblock {Moral Decision Making in Autonomous Systems: Enforcement, Moral
  Emotions, Dignity, Trust, and Deception}.
\newblock {\em Proceedings of the IEEE} 100(3):571--589.

\bibitem[\protect\citeauthoryear{Caminada \bgroup et al\mbox.\egroup
  }{2014}]{CaminadaKOV14}
Caminada, M. W.~A.; Kutl{\'{a}}k, R.; Oren, N.; and Vasconcelos, W.~W.
\newblock 2014.
\newblock {Scrutable Plan Enactment via Argumentation and Natural Language
  Generation}.
\newblock In {\em Proc. International conference on Autonomous Agents and
  Multi-Agent Systems (AAMAS)},  1625--1626.
\newblock {IFAAMAS/ACM}.

\bibitem[\protect\citeauthoryear{Clarke, Grumberg, and Peled}{1999}]{Clarke99}
Clarke, E.; Grumberg, O.; and Peled, D.
\newblock 1999.
\newblock {\em {Model Checking}}.
\newblock MIT Press.

\bibitem[\protect\citeauthoryear{Dennis and Fisher}{2017}]{DennisFisher18:AIES}
Dennis, L.~A., and Fisher, M.
\newblock 2017.
\newblock {Arbiters of Acceptable Behaviour}.
\newblock Under review.

\bibitem[\protect\citeauthoryear{Dennis \bgroup et al\mbox.\egroup
  }{2016a}]{Dennis16}
Dennis, L.; Fisher, M.; Slavkovik, M.; and Webster, M.
\newblock 2016a.
\newblock {Formal Verification of Ethical Choices in Autonomous Systems}.
\newblock {\em Robotics and Autonomous Systems} 77:1--14.

\bibitem[\protect\citeauthoryear{Dennis \bgroup et al\mbox.\egroup
  }{2016b}]{Dennis14}
Dennis, L.~A.; Fisher, M.; Lincoln, N.~K.; Lisitsa, A.; and Veres, S.~M.
\newblock 2016b.
\newblock {Practical Verification of Decision-Making in Agent-Based Autonomous
  Systems}.
\newblock {\em Automated Software Engineering} 23(3):305--359.

\bibitem[\protect\citeauthoryear{Dennis, Fisher, and Winfield}{2015}]{Dennis15}
Dennis, L.~A.; Fisher, M.; and Winfield, A.
\newblock 2015.
\newblock {Towards Verifiably Ethical Robot Behaviour}.
\newblock In {\em Proceedings of the AAAI Workshop on Artificial Intelligence
  and Ethics (1st International Workshop on AI and Ethics)}.
\newblock IEEE Press.

\bibitem[\protect\citeauthoryear{Dignum \bgroup et al\mbox.\egroup
  }{2017}]{Dignum18}
Dignum, V.; Baldoni, M.; Baroglio, C.; Caon, M.; Chatila, R.; Dennis, L.;
  G\'{e}nova, G.; Klie\ss, M.; Lopez-Sanches, M.; Micalizio, R.; Pav\'{o}n, J.;
  Slavkovik, M.; Smakman, M.; van Steenbergen, M.; Tedeschi, S.; van~der Torre,
  L.; Villata, S.; de~Wildt, T.; and Haim, G.
\newblock 2017.
\newblock {Ethics by Design: necessity or curse?}
\newblock Under Review.

\bibitem[\protect\citeauthoryear{Miettinen}{1998}]{Miettinen98}
Miettinen, K.
\newblock 1998.
\newblock {\em Nonlinear Multiobjective Optimization}, volume~12 of {\em
  International Series in Operations Research \&{} Management Science}.
\newblock Springer.

\bibitem[\protect\citeauthoryear{Moor}{2006}]{Moor06}
Moor, J.~H.
\newblock 2006.
\newblock {The Nature, Importance, and Difficulty of Machine Ethics}.
\newblock {\em IEEE Intelligent Systems} 21(4):18--21.

\bibitem[\protect\citeauthoryear{Rao and Georgeff}{1995}]{Rao95}
Rao, A.~S., and Georgeff, M.~P.
\newblock 1995.
\newblock {BDI Agents: From Theory to Practice}.
\newblock {\em Proceedings of the First International Conference on Multiagent
  Systems} 95:312--319.

\bibitem[\protect\citeauthoryear{Shim and Arkin}{2017}]{Shim17}
Shim, J., and Arkin, R.~C.
\newblock 2017.
\newblock {An Intervening Ethical Governor for a Robot Mediator in
  Patient-Caregiver Relationships}.
\newblock In Aldinhas~Ferreira, M.~I.; Silva~Sequeira, J.; Tokhi, M.~O.;
  E.~Kadar, E.; and Virk, G.~S., eds., {\em A World with Robots: International
  Conference on Robot Ethics: ICRE 2015},  77--91.
\newblock Springer International Publishing.

\bibitem[\protect\citeauthoryear{Vanderelst and Winfield}{2016}]{Vanderelst16}
Vanderelst, D., and Winfield, A.
\newblock 2016.
\newblock {An Architecture for Ethical Robots inspired by the Simulation Theory
  of Cognition}.
\newblock {\em Cognitive Systems Research}.

\bibitem[\protect\citeauthoryear{Winfield and Jirotka}{2017}]{Winfield17}
Winfield, A., and Jirotka, M.
\newblock 2017.
\newblock The case for an ethical black box.
\newblock In {\em Towards Autonomous Robotic Systems}.
\newblock Springer.

\bibitem[\protect\citeauthoryear{Winfield, Blum, and Liu}{2014}]{Winfield13}
Winfield, A. F.~T.; Blum, C.; and Liu, W.
\newblock 2014.
\newblock {Towards and Ethical Robot: Internal Models, Consequences and Ethical
  Action Selection}.
\newblock In Mistry, M.; Leonardis, A.; Witkowski, M.; and Melhuish, C., eds.,
  {\em Advances in Autonomous Robotics Systems}, volume 8717 of {\em Lecture
  Notes in Computer Science},  85--96.
\newblock Springer.

\end{thebibliography}
\end{document}